\documentclass[runningheads]{llncs}
\usepackage{amsmath}
\usepackage{amssymb}
\usepackage{mathtools}

\usepackage{rotating}
\usepackage{xcolor}
\usepackage{soul}
\usepackage{colortbl}
\usepackage{booktabs}
\usepackage{relsize}
\usepackage{tabularx}
\usepackage{setspace}
\usepackage{float}
\usepackage{supertabular}

\usepackage{subcaption}

\usepackage{hyperref}

\newcommand{\etal}{\textit{et al.\ }}

\begin{document}
\title{MEGAN: Multi-Explanation Graph Attention Network}

\newif\ifanon

\ifanon
\author{Anonymous}
\authorrunning{Anon et al.}
\institute{
Anonymous Institute\\
\email{\{abc,lncs\}@anon.edu}}

\else
\author{
Jonas Teufel\inst{1}\orcidID{0000-0002-9228-9395} \and
Luca Torresi\inst{1}\orcidID{0000-0003-2205-6753} \and
Patrick Reiser\inst{1}\orcidID{0000-0002-7052-696X} \and
Pascal Friederich\thanks{corresponding author}\inst{1}\orcidID{0000-0003-4465-1465}}
\authorrunning{
J. Teufel et al.
}
\institute{
Institute of Theoretical Informatics (ITI), Karlsruhe Institute of Technology (KIT), Karlsruhe, Germany\\
\email{\{jonas.teufel\}@student.kit.edu}\\
\email{\{luca.torresi,patrick.reiser,pascal.friederich\}@kit.edu}}
\fi

\maketitle              

\begin{abstract}
We propose a multi-explanation graph attention network (MEGAN). Unlike existing graph explainability methods, our network can produce node and edge attributional explanations along multiple channels, the number of which is independent of task specifications. This proves crucial to improve the interpretability of graph regression predictions, as explanations can be split into positive and negative evidence w.r.t to a reference value. Additionally, our attention-based network is fully differentiable and explanations can actively be trained in an explanation-supervised manner. We first validate our model on a synthetic graph regression dataset with known ground-truth explanations. Our network outperforms existing baseline explainability methods for the single- as well as the multi-explanation case, achieving near-perfect explanation accuracy during explanation supervision. Finally, we demonstrate our model's capabilities on multiple real-world datasets. We find that our model produces sparse high-fidelity explanations consistent with human intuition about those tasks.
\end{abstract}



\section{Introduction}

Explainable AI (XAI) methods aim to provide explanations complementing a model's predictions to make it's complex inner workings more transparent to humans with the intention to improve trust and reliability, provide tools for model analysis, and comply with anti-discrimination laws \cite{doshi-velez_towards_2017}. The majority of existing work on graph explainability focuses on post-hoc methods, which can be used to generate explanations for already trained models, which have been proven to perform well. While post-hoc methods are an important area of development to add explainability to time-tested models, we want to emphasize the potential of \textit{self-explaining} methods. In their literature review, Jiminez-Luna \etal \cite{jimenez-luna_drug_2020} describe these methods as being explainable by design. One example of this class are the simpler, traditional machine learning approaches that are naturally interpretable, such as decision tree methods \cite{friederich_scientific_2021}. However, we want to focus on self-explaining graph neural networks, which produce the attributional explanations for the nodes and edges of the input graph directly alongside each prediction. We emphasize this class of methods specifically due to their capability for explanation-supervised training. During explanation-supervised training, a model is additionally trained to produce explanations that are similar to a given set of reference explanations. Recently, there has been promising progress on the topic of explanation supervision in the domains of image processing \cite{linsley_learning_2019,qiao_exploring_2018,boyd_cyborg_2022} and natural language processing \cite{fernandes_learning_2022,pruthi_learning_2020,stacey_supervising_2022}. Previous work is able to improve model interpretability by training models to generate more human-like explanations and even improve main prediction performance by training models on human-generated image saliency maps. In the graph domain, however, there has been little work on explanation supervision \cite{gao_gnes_2021,magister_encoding_2022} yet. Inspired by the successes recently demonstrated in other domains, we propose the self-explaining \textit{multi-explanation graph attention network} (MEGAN) architecture. In this work, we demonstrate that our model shows significantly improved capability to learn explanations during explanation-supervised training, outperforming the baseline method \cite{gao_gnes_2021} from the literature.\\

\noindent In addition to its properties w.r.t. explanation supervision, we design our network to output explanations along \textit{multiple channels}, the number of which is independent of the main prediction task. Like the majority of existing GNN explainability methods, we focus on attributional explanations, which attribute a value of importance to each element of the input graph. For existing methods, the number of these attribution values is dictated by the details of the main prediction task. For single-value graph regression tasks for example a single value would be assigned to each node and edge. For our multi-explanation method, however, this number of attributions is a property of the network rather than restricted by task specifications.\\
We want to emphasize the importance of this property especially in regard to graph regression problems. For the prediction of a single regression value, existing methods only produce a single attribution for each node and edge. We argue that such explanations are insufficient for the interpretation of regression predictions. In reality, one often encounters structure-property explanations of opposing \textit{polarity}. One practical example of this is the prediction of water solubility, where large non-polar carbon structures generally cause low solubility values and polar functional groups cause higher values. A single attributional explanation may highlight all the important motifs, but is not able to capture this crucial detail about their polarity. For this reason, we decouple the number of explanations from the task specification to be able to produce two explanations (negative and positive influence) for graph regression problems. We introduce an explanation co-training method which uses only the generated explanation masks to solve an approximation of the prediction problem to promote each explanation channel to behave according to their intended interpretation. In our experiments, we find that this explanation co-training is an effective method to guide the generation of the explanation channels to contribute faithfully to the prediction outcome according to pre-determined interpretations. We validate this finding on several real-world datasets, where our model produces explanations consistent with human intuition about those tasks. Beyond that, we apply our model to one real-world task of molecular property prediction without common human intuition and are able to support previously published hypotheses about structure-property relationships and propose several new potential explanatory motifs.

\section{Related Work}


\subsubsection{GNN Explanation Methods} 

Yuan \etal \cite{yuan_explainability_2022} provide a taxonomic overview of XAI methods for graph neural networks. Some methods have been adapted from similar approaches in other domains, such as GradCAM \cite{pope_explainability_2019}, GraphLIME \cite{huang_graphlime_2022} and LRP \cite{schwarzenberg_layerwise_2019}. Other methods were developed specifically for graph neural networks. Notable ones include GNNExplainer \cite{ying_gnnexplainer_2019}, PGExplainer \cite{luo_parameterized_2020}, and Zorro \cite{funke_zorro_2023}. Jiminez-Luna \etal \cite{jimenez-luna_drug_2020} present another literature review about the applications of XAI in drug discovery. Henderson \etal \cite{henderson_improving_2021} for example introduce regularization terms to improve GradCAM-generated explanations for chemical property prediction. Sanchez-Lenglin \etal \cite{sanchez-lengeling_evaluating_2020} introduce new benchmark datasets for attributional graph explanations based on molecular graphs and compare several existing explanation methods.\\
Generally, most explanation methods aim to produce attributional explanations, which explain a prediction by assigning importance values to the nodes and edges of the input graph. However, there exists some criticism about this class of explanations \cite{adebayo_sanity_2018,kindermans_reliability_2019}, which is partially why recently different modalities of explanations have been explored for the graph domain as well. Magister \etal \cite{magister_gcexplainer_2021} for example propose GCExplainer, which can be used to generate \textit{concept-based} explanations for graph neural networks in a post-hoc fashion. Shin \etal \cite{shin_page_2022} for example propose PAGE, a method to generate \textit{prototype-based} explanations. \textit{Counterfactuals} are yet another popular explanation modality, for which Tan \etal \cite{tan_learning_2022} and Prado-Romero and Stilo \cite{prado-romero_gretel_2022} have recently proposed methods for graph neural networks.

\subsubsection{Self-Explaining Graph Neural Networks}

In their literature review, Jiminez-Luna \etal \cite{jimenez-luna_drug_2020} define \textit{self-explaining} methods as those that are explainable by design. One large fraction of this category is represented by simpler traditional machine learning methods. Friederich \etal \cite{friederich_scientific_2021} for example use an interpretable decision tree approach to structure-property relationships for several real-world graph datasets. However, there is also recent progress for more complex self-explaining models such as graph neural networks. Dang and Wang \cite{dai_towards_2021} and Zhang \etal \cite{zhang_protgnn_2022} independently introduce self-explaining graph neural networks for prototype-based explanations. Magister \etal \cite{magister_encoding_2022} introduce a self-explaining network for concept-based explanations. Furthermore, Müller \etal \cite{muller_dtgnn_2022} propose DT+GNN, an interesting method that combines the capabilities of GNNs with the inherent interpretability of decision trees.

\subsubsection{Explanation Supervision}

During explanation supervision, models are not only trained to perform a main prediction task through ground truth target labels but also to produce explanations that are similar to a given set of reference explanations. Most interestingly explanation supervision provides the possibility to train models to produce more human-like explanations. Beyond that, several works are able to show that the inclusion of human saliency maps has the potential to increase the task performance of the models \cite{linsley_learning_2019,boyd_cyborg_2022}. In that context, Linseley \etal \cite{linsley_learning_2019} for example show that human saliency maps improve the performance of an image classifier. Boyd et al. \cite{boyd_cyborg_2022} demonstrate that human saliency annotations improve the performance of a deep fake detection model. In the domain of natural language processing, Pruthi \etal \cite{pruthi_learning_2020} use explanation-supervised models to substitute human participants in artificial simulatability studies to assess the quality of explanations. Fernandes \etal \cite{fernandes_learning_2022} even take this concept one step further and train an explainer to optimize this property of simulatability.

\section{Multi-Explanation Graph Attention Network}


\subsection{Task Description}
\label{sec:task}


We assume a directed graph $\mathcal{G} = (\mathcal{V}, \mathcal{E})$ is represented by a set of node indices $\mathcal{V} \subset \mathbb{N}^{V}$ and a set of edges $\mathcal{E} \subseteq \mathcal{V} \times \mathcal{V} \subset \mathbb{R}^{E}$, where a tuple $(i, j) \in \mathcal{E}$ denotes an edge from node $i$ to node $j$. Every node $i$ is associated with a vector of initial node features $\mathbf{h}_{i}^{\text{(0)}} \in \mathbb{R}^{N_0}$, combining into the initial node feature tensor $\mathbf{H}^{\text{(0)}} \in \mathbb{R}^{V \times N_0}$. Each edge is associated with a feature vector $\mathbf{u}_i \in \mathbb{R}^{M}$, combining into the edge feature tensor $\mathbf{U} \in \mathbb{R}^{E \times M}$.\\
We consider graph classification and regression problems, which means graphs are associated with a target vector $\mathbf{y} \in \mathbb{R}^{C}$ which is either a one-hot class encoding or continuous regression values. In addition, node and edge attributional explanations for graphs are considered. We define explanations as masks that assign $[0, 1]$ values to each node and each edge, representing the importance of the corresponding graph element toward the outcome of the prediction. We generally assume that any prediction may be explained by $K$ individual importance channels, where $K$ is an independent hyperparameter. The node explanations are given as the \textit{node importance} tensor $\mathbf{V}^{\text{im}} \in [0,1]^{V \times K}$ and the edge explanations are given as the \textit{edge importance} tensor $\mathbf{E}^{\text{im}} \in [0,1]^{E \times K}$.


\subsection{Architecture Overview}
\label{sec:architecture}

%

\begin{figure}[t]
\begin{center}
\includegraphics[width=1.0\textwidth]{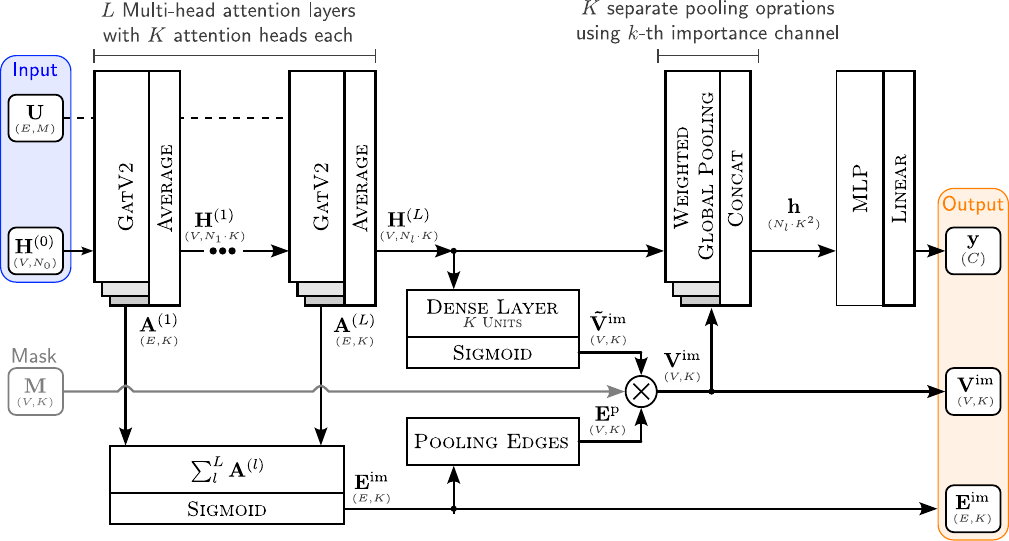}
\end{center}
\caption{Multi-explanation graph attention network (MEGAN) architecture overview. Rectangle boxes represent layers; arrows indicate layer interconnections. Rounded boxes represent tensors. Intermediate tensors are also named annotated arrows. Tuples beneath variable names indicate the tensor shape, with batch dimension omitted, but implicitly assumed as the first dimension for all.}
\label{fig:architecture}
\end{figure}

\begin{figure}[t]
\begin{center}
\includegraphics[width=1.0\textwidth]{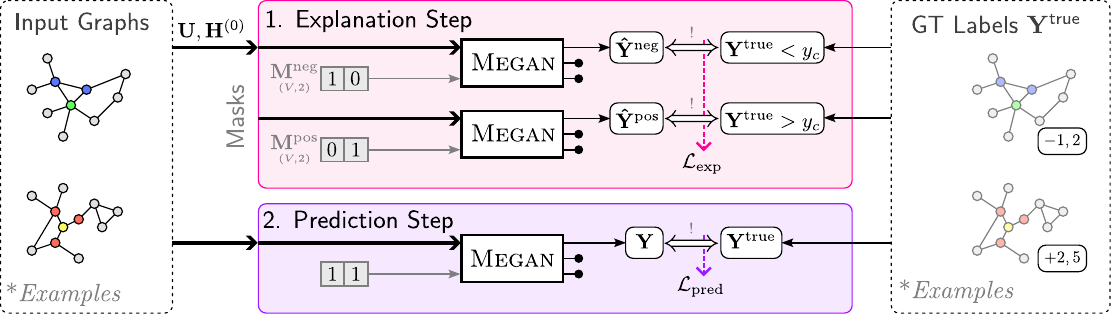}
\end{center}
\caption{Illustration of the split training procedure for the regression case. The explanation-only train step attempts to find an approximate solution to the main prediction task, by using only a globally pooled node importance tensor. After the weight update for the explanation step was applied to the model, the prediction step performs another weight update based on the actual output of the model and the ground truth labels.}
\label{fig:train-step}
\end{figure}

To solve the previously defined task we propose the following \textit{multi-explanation graph attention network} (MEGAN) architecture, for which Figure \ref{fig:architecture} provides a visual overview. The network consists of $L$ attention layers, where the number of layers $L$ and the hidden units of each layer are hyperparameters. Each of these layers consists of $K$ individual, yet structurally identical GATv2 \cite{brody_how_2022} attention heads, one for each of the $K$ expected explanation channels. Assuming the attention heads in the $l$-th layer have $N_l$ hidden units, then each attention head produces its own node embeddings $\mathbf{H}^{(l,k)}$, where $k \in \{1,\dots,K\}$ is the head index. The final node embeddings $\mathbf{H}^{(l)} \in \mathbb{R}^{V \times N_l \cdot K}$
of layer $l$ are then produced by averaging all these individual matrices along the feature dimension: 
\begin{equation}
\mathbf{H}^{(l)} = \frac{1}{K} \sum^{K}_k \mathbf{H}^{(l,k)}
\end{equation}
This node embedding tensor is then used as the input to \textit{each} of the $K$ attention heads of layer $l+1$. 
Aside from the node embeddings, each attention head also produces a vector $\mathbf{A}^{(l,k)} \in \mathbb{R}^{E}$ of attention logits which are used to calculate the attention weights 
\begin{equation}
{\pmb \alpha}^{(l,k)} = \operatorname{softmax} (\mathbf{A}^{(l,k)})
\end{equation}
of the $k$-the attention head in the $l$-th layer. 
The edge importance tensor $\mathbf{E}^{\text{im}} \in [0,1]^{E \times K}$ is calculated from the concatenation of these attention logit tensors in the feature dimension and summed up over the number of layers:
\begin{equation}
\mathbf{E}^{\text{im}} = \sigma \left( \sum_{l=1}^{L} \left( \mathbf{A}^{(l,1)} \;||\; \mathbf{A}^{(l,2)}  \;||\; \dots  \;||\; \mathbf{A}^{(l,K)} \right) \right)
\end{equation}
Based on this, a local pooling operation is used to derive the pooled edge importance tensor $\mathbf{E}^{\text{p}} \in [0, 1]^{V \times K}$ for the \textit{nodes} of the graph. This local pooling operation can be seen as the aggregation step in a message-passing framework, where the edge importance values are treated as the corresponding messages.\\
The final node embeddings $\mathbf{H}^{(L)}$ are then used as the input to a dense network, whose final layer is set to have $K$ hidden units, producing the node importance embeddings $\tilde{\mathbf{V}}^{\text{im}} \in [0, 1]^{V \times K}$. The node importance tensor is then calculated as the product of those node importance embeddings $\tilde{\mathbf{V}}^{\text{im}} \in [0, 1]^{V \times K}$ and the pooled edge importance tensor $\mathbf{E}^{\text{p}} \in [0, 1]^{V \times K}$:
\begin{equation}
\mathbf{V}^{\text{im}} = \tilde{\mathbf{V}}^{\text{im}} \cdot \mathbf{E}^{\text{p}} \cdot \mathbf{M}.
\end{equation}
The mask $\mathbf{M}$ introduced in Fig.~\ref{fig:architecture} is only optionally used to compute the fidelity metric, which is introduced in Section~\ref{sec:fidelity}.
At this point, the edge and node importance matrices, which represent the explanations generated by the network, are already accounted for, which leaves only the primary prediction to be explained. The first remaining step is a global sum pooling operation which turns the node embedding tensor $\mathbf{H}^{(L)}$ into a vector of global graph embeddings. For this, $K$ separate weighted global sum pooling operations are performed, one for each explanation channel. Each of these pooling operations uses the same node embeddings $\mathbf{H}^{(L)}$ as input, but a different slice $V^{\text{im}}_{:,k}$ of the node importance tensor as weights. In that way, $K$ separate graph embedding vectors 
\begin{equation}
\mathbf{h}^{(k)} = \sum^{V}_{i=0} \left( \mathbf{H}^{(L)} \cdot \mathbf{V}^{\text{im}}_{:,k} \right)_{i,:}
\end{equation}
are created, which are then concatenated into a single graph embedding vector 
\begin{equation}
\mathbf{h} = \mathbf{h}^{(1)} \: || \: \mathbf{h}^{(2)} \: || \: \dots \: || \: \mathbf{h}^{(K)}
\end{equation}
where $\mathbf{h} \in \mathbb{R}^{N_L \cdot K}$. This graph embedding vector is then passed through a generic MLP whose final layer either has linear activation for graph regression or softmax activation for graph classification to create an appropriate output
\begin{equation}
\mathbf{y} = \text{MLP} (\mathbf{h})
\end{equation}


\subsection{Explanation Co-Training}
\label{sec:explanation-training}
With the architecture as explained up to this point, there is no mechanism yet to ensure that individual explanation channels learn the appropriate explanations according to their intended interpretation (for example positive vs negative evidence). We use a special explanation co-training procedure to guide the individual explanation channels to develop according to pre-determined interpretations. This is illustrated in Figure \ref{fig:train-step}. For this purpose, the loss function consists of two parts: The prediction loss and the explanation loss. The explanation loss is based only on the node importance tensor produced by the network. A global sum pooling operation is used to turn the importance values of each separate channel into a single \textit{alternate output tensor} $\hat{\mathbf{Y}} \in \mathbb{R}^{B \times K}$, where $B$ is the training batch size. This alternate output tensor is then used to solve an approximation of the original prediction problem: This can be seen as a reduction of the problem into a set of $K$ separate and independent subgraph counting problems, where each of those only uses the subset of training batch samples that aligns with the respective channel's intended interpretation.
 
\subsubsection{Regression} For regression, we assume $K = 2$, where the first channel represents the negative and the second channel the positive influences relative to the reference value $y_c$, which is a hyperparameter of the model and usually set as the arithmetic mean of the target value distribution in the train set.
We select all samples of the current training batch lesser and greater than the reference value and use these to calculate a mean squared error (MSE) loss:
\begin{equation}
\mathcal{L}_{\text{exp}} = \frac{1}{2 \cdot B} \sum_{b=1}^{B} \begin{cases}
(\mathbf{\hat{Y}}_{b,0} - y_c - \mathbf{Y}^{\text{true}}_{b})^2 & \text{if } \mathbf{Y}^{\text{true}}_{b} < y_c \\
(\mathbf{\hat{Y}}_{b,1} - y_c - \mathbf{Y}^{\text{true}}_{b})^2 & \text{if } \mathbf{Y}^{\text{true}}_{b} > y_c \\
\end{cases}
\end{equation}

\subsubsection{Classification} We assume the number of channels $K = C$ is equal to the number of possible output classes $C$. We use the alternate output channel to compute an individual binary cross entropy (BCE) loss for each channel:
\begin{equation}
\mathcal{L}_{\text{exp}} = \frac{1}{C \cdot B} \sum^B_{b=1} \sum^C_{c=1} \mathcal{L}_{\text{BCE}}( \mathbf{Y}^{\text{true}}_{b,c}, \mathbf{\hat{Y}}_{b,c})
\end{equation}

\noindent For regression as well as classification, the total loss during model training consists of these task-specific terms and an additional term for explanation sparsity: 
\begin{equation}
\mathcal{L}_{\text{total}} = \mathcal{L_{\text{pred}}} + \gamma \mathcal{L}_{\text{exp}} + \beta \mathcal{L}_{\text{sparsity}}
\end{equation}
where $\gamma$ and $\beta$ are hyperparameters of the training process. Explanation sparsity $\mathcal{L}_{\text{sparsity}}$ is calculated as L1 regularization over the node importance tensor. Based on this loss the gradients are calculated and the model weights are updated.\\

\noindent We will henceforth use the notation $\text{MEGAN}^{K}_{\gamma}$ to refer to specific model configurations with $K$ explanation channels, $\gamma$ explanation co-training weight and use the superscript $\text{MEGAN}^{\text{(S)}}$ to indicate when models where trained in an explanation-supervised fashion.


\subsection{Multi-Channel Fidelity}
\label{sec:fidelity}

A particular challenge in the field of explainable AI is the question of how to properly assess the quality of explanations \cite{doshi-velez_towards_2017}. One commonly used metric is the \textit{fidelity} of explanations w.r.t. the model predictions. It quantifies the extent to which the explanation is responsible for the corresponding prediction. Yuan \etal \cite{yuan_explainability_2022} define the $\operatorname{Fidelity}^+$ metric as the deviation of the predicted model output if all the nodes and edges that are part of the explanation are removed from the input. The reasoning is that the higher this resulting output deviation, the more important the explanation must have been for the original prediction. This metric is usually computed by setting all the features of the corresponding nodes and edges of the input graph to zero. However, one issue with this approach is that zero might be an in-distribution value for the input features. Therefore, the masked input elements may have an effect on the model that is different than their intended removal.\\

\noindent To address this issue we introduce the multi-channel $\operatorname{Fidelity}^*$ metric to assess the faithfulness of MEGAN's predictions. Since our network directly incorporates the explanations into the prediction process as weights of the final global pooling operation, we can directly manipulate these explanations to quantify their impact on the prediction. This can be done by providing an additional importance mask $\mathbf{M} \in [0, 1]^{V \times K}$ during the prediction of the network (see Figure~\ref{fig:architecture}). For each explanation channel $k$, we construct a mask $\mathbf{M}^k$ which only suppresses that channel from the final pooling operation. The model is then queried with that mask to produce the modified output $\mathbf{\hat{y}}^k$, which we use to calculate the deviation $\Delta^k = |\mathbf{y} - \mathbf{\hat{y}}_k|$ w.r.t. the original output. The fidelity is then calculated as:
\begin{equation}
    \operatorname{Fidelity}^* = \frac{1}{K} \sum_{k}^{K} \begin{cases}
        +\Delta^k & \text{if deviation as expected for channel } k \\
        -\Delta^k & \text{if deviation }not\text{ as expected for channel } k
    \end{cases}
\end{equation}
\noindent What kind of deviation counts as \textit{expected} for a given channel $k$ is defined by the interpretation that is assigned to that channel. In the case of regression, for example, we assign the interpretation of the first explanation channel to be the negatively influencing evidence and the second channel to be the positively influencing evidence. In that case, if all the negative evidence is omitted from the result, it would be expected that the output becomes more positive than the original prediction and vice versa. For classification on the other hand, if all evidence for one specific class is suppressed it would be expected that the confidence of that respective class decreases.\\
Consequently, a positive $\operatorname{Fidelity}^*$ indicates that the channels of the model generally have an effect on the prediction outcome that matches with their pre-defined interpretation.
\section{Computational Experiments}
\label{sec:experiments}

We conduct computational experiments to demonstrate the capabilities of our network. Primarily, we emphasize two key strengths of our proposed model: (1) The inherent advantage of multi channel-explanations especially in regard to the interpretability of regression problems. On a specifically designed synthetic dataset we show that, unlike other post-hoc methods, by using explanation co-training our model is able to correctly capture the \textit{polarity} of existing sub-graph evidence. (2) Our model's significantly increased capability for explanation-supervised training, where our model correctly learns to replicate the ground truth explanations that it was trained on. Additionally, we conduct experiments with real-world graph classification and regression datasets that provide anecdotal evidence for the correctness of the model's explanations for more complex tasks as well.

\begin{figure}[t]
\begin{center}
\includegraphics[width=1.0\textwidth]{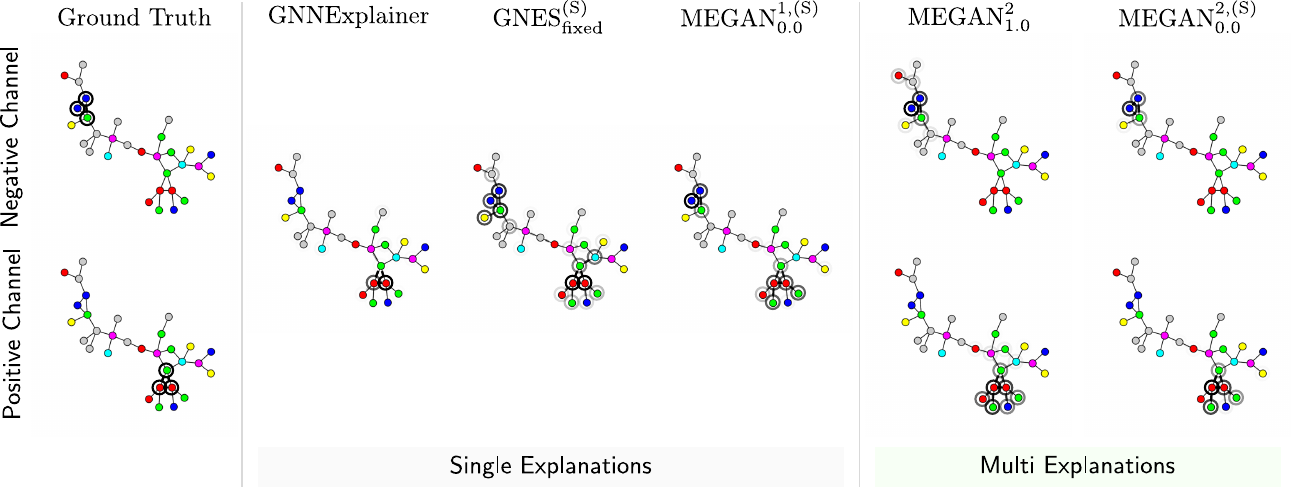}
\end{center}
\caption{Examples for explanations generated for one element of the RbMotifs dataset using selected methods. Explanations are represented as bold highlights of the corresponding graph elements. Left: The ground truth explanations split by the polarity of their influence on the graph target value. Middle: Explanations generated by some selected single-explanation methods. Right: Explanations generated by the multi-explanation MEGAN models.}
\label{fig:rbmotifs}
\end{figure}


\subsection{Synthetic Graph Regression}

We create a synthetic graph regression dataset called \textit{RbMotifs} consisting of 5000 randomly generated graphs, where each node is associated with 3 node features representing an RGB color code. Graphs are additionally randomly seeded with specific simple sub-graph motifs, which either consist dominantly of red nodes or blue nodes. If a red-based motif exists within a graph, it contributes a constant positive value to the overall target value of a graph. Likewise, a blue-based motif contributes a negative value. Thus, the overall target value associated with each graph is the sum of all the sub-graph contributions and a small random value. The dataset  represents a simple motif-based graph regression problem, where the individual sub-graph motifs are considered the perfect ground truth explanations. Most importantly, the explanations have a clear \textit{opposing polarity} which is crucial to the understanding of the dataset's underlying structure-property relationship.


\subsubsection{Single Explanations}

Although many regression tasks may exhibit such explanations of different polarity, existing post-hoc attributional XAI methods are only able to provide a single explanation. These single explanations are only able to point out which parts of the graph are generally important for the prediction but do not capture in what manner they contribute to the outcome. Therefore, to compare our proposed MEGAN model to some established existing post-hoc explanation methods, we conduct a first experiment that only considers such single explanations. For this case, we concatenate all of the relevant sub-graphs into a single channel which will be considered the ground truth explanation for each element of the dataset.\\

\noindent We conduct the experiment for explanations obtained from Gradients \cite{pope_explainability_2019},  GNNExplainer \cite{ying_gnnexplainer_2019}, GNES \cite{gao_gnes_2021} and MEGAN. For all the post-hoc methods we train a 3-layer GATv2 network as the basis for the explanations. The results of this experiment can be found in Table~\ref{tab:rbmotifs_all}. We report on the overall prediction performance of the network, the explanation accuracy, the sparsity, and the fidelity of the explanations. The explanation accuracy is given as the node and edge AUROC score resulting from a comparison with the ground truth explanations, as it is proposed by McCloskey \etal \cite{mccloskey_using_2019}. The fidelity is given as the relative value $\operatorname{Fidelity}^+_{\text{rel}}$, which is the difference between the predicted explanation's fidelity and the fidelity of random explanations of the same sparsity (see Appendix~\ref{app:fidelity}). In addition, we perform experiments with explanation supervision. To our knowledge, MEGAN and GNES are currently the only methods capable of explanation supervision for node and edge attributional explanations. For both of these cases, the models are trained with ground truth explanations in addition to the target values.\\

\begin{table}[h!]
\caption{Results for 25 independent repetitions of the computational experiments on the RbMotifs dataset. We report the mean in black and the standard deviation in gray. The upper section contains results for the single-explanation experiments and the lower section for multi-explanation experiments. We highlight the best results in each section in bold and underline the second-best.}
\label{tab:rbmotifs_all}
\renewcommand*{\arraystretch}{1.6}
\setlength\tabcolsep{5pt}
\begin{center}
\smaller

\definecolor{LightGreen}{rgb}{0.97,1,0.97}
\newcolumntype{a}{>{\columncolor{LightGreen}}c}
\begin{tabular}{lcaacc}
\toprule
\multicolumn{1}{c}{Explanations} &
\multicolumn{1}{c}{$\text{r}^2 \uparrow$} &
\multicolumn{1}{a}{Node AUC $\uparrow$} &
\multicolumn{1}{a}{Edge AUC $\uparrow$} &
\multicolumn{1}{c}{Sparsity $\downarrow$} &
\multicolumn{1}{c}{$\operatorname{Fidelity}^{+}_{\text{rel}}$ $\uparrow$}\\
\midrule


$\text{Gradients}$  & 
$0.89{ \color{gray} \mathsmaller{ \pm 0.05 } } $  & 
$0.73{ \color{gray} \mathsmaller{ \pm 0.05 } } $  & 
$0.60{ \color{gray} \mathsmaller{ \pm 0.03 } } $  & 
$0.12{ \color{gray} \mathsmaller{ \pm 0.01 } } $  & 
$0.57{ \color{gray} \mathsmaller{ \pm 0.14 } } $  \\ 


$\text{GnnExplainer}$  & 
$0.89{ \color{gray} \mathsmaller{ \pm 0.05 } } $  & 
$0.70{ \color{gray} \mathsmaller{ \pm 0.04 } } $  & 
$0.52{ \color{gray} \mathsmaller{ \pm 0.03 } } $  & 
$0.22{ \color{gray} \mathsmaller{ \pm 0.06 } } $  & 
$\underline{0.78}{ \color{gray} \mathsmaller{ \pm 0.20 } } $  \\

$\text{GNES}_{\text{original}}^{\text{(S)}}$  & 
$0.88{ \color{gray} \mathsmaller{ \pm 0.02 } } $  & 
$0.63{ \color{gray} \mathsmaller{ \pm 0.04 } } $  & 
$0.58{ \color{gray} \mathsmaller{ \pm 0.03 } } $  & 
$0.10{ \color{gray} \mathsmaller{ \pm 0.01 } } $  & 
$0.50{ \color{gray} \mathsmaller{ \pm 0.22 } } $  \\ 

$\text{GNES}_{\text{fixed}}^{\text{(S)}}$  & 
$0.88{ \color{gray} \mathsmaller{ \pm 0.02 } } $  & 
$\underline{0.85}{ \color{gray} \mathsmaller{ \pm 0.04 } } $  & 
$0.66{ \color{gray} \mathsmaller{ \pm 0.02 } } $  & 
$0.12{ \color{gray} \mathsmaller{ \pm 0.01 } } $  & 
$0.74{ \color{gray} \mathsmaller{ \pm 0.13 } } $  \\ 

$\text{MEGAN}_{0.0}^{1}$  & 
$\underline{0.92}{ \color{gray} \mathsmaller{ \pm 0.05 } } $  & 
$0.82{ \color{gray} \mathsmaller{ \pm 0.12 } } $  & 
$\underline{0.79}{ \color{gray} \mathsmaller{ \pm 0.08 } } $  & 
$0.14{ \color{gray} \mathsmaller{ \pm 0.08 } } $  & 
$\mathbf{1.10}{ \color{gray} \mathsmaller{ \pm 3.03 } } $  \\ 

$\text{MEGAN}_{0.0}^{1, \text{(S)}}$   & 
$\mathbf{0.95}{ \color{gray} \mathsmaller{ \pm 0.02 } } $  & 
$\mathbf{0.98}{ \color{gray} \mathsmaller{ \pm 0.00 } } $  & 
$\mathbf{0.99}{ \color{gray} \mathsmaller{ \pm 0.00 } } $  & 
$0.18{ \color{gray} \mathsmaller{ \pm 0.00 } } $  & 
$0.53{ \color{gray} \mathsmaller{ \pm 0.17 } } $ \\

\midrule


$\text{MEGAN}^{2}_{1.0}$  & 
$\underline{0.95}{ \color{gray} \mathsmaller{ \pm 0.01 } } $  & 
$\underline{0.94}{ \color{gray} \mathsmaller{ \pm 0.02 } } $  & 
$\underline{0.85}{ \color{gray} \mathsmaller{ \pm 0.06 } } $  & 
$0.10{ \color{gray} \mathsmaller{ \pm 0.06 } } $  & 
$\underline{2.06}^{\text{(*)}}{ \color{gray} \mathsmaller{ \pm 0.85 } } $  \\ 

$\text{MEGAN}^{2, \text{(S)}}_{0.0}$  & 
$\mathbf{0.95}{ \color{gray} \mathsmaller{ \pm 0.03 } } $  & 
$\mathbf{0.99}{ \color{gray} \mathsmaller{ \pm 0.00 } } $  & 
$\mathbf{0.99}{ \color{gray} \mathsmaller{ \pm 0.00 } } $  & 
$0.09{ \color{gray} \mathsmaller{ \pm 0.06 } } $  & 
$\mathbf{2.11}^{\text{(*)}}{ \color{gray} \mathsmaller{ \pm 0.36 } } $ \\ 

\bottomrule
\end{tabular}
\end{center}
\smaller \smaller
\hspace*{2pt} $^{\text{(S)}}$ Explanation-supervised models. These models were trained on the ground truth explanation annotations in addition to the main target values.\\
\hspace*{2pt} $^{\text{(*)}}$ Values of the multi-channel $\operatorname{Fidelity}^*$ metric. Note that these are \textit{not} comparable to the other fidelity values obtained in a single channel setting. 
\end{table}

\noindent The results show that the explanations generated by all the methods achieve reasonable results for predictive performance, the node accuracy w.r.t. the explanation ground truth, as well as sparsity and fidelity. The explanation supervised methods show the best results for explanation accuracy. The supervised $\text{MEGAN}^{1,\text{(S)}}_{0.0}$ model achieves a near-perfect accuracy, with the explanation-supervised GNES method being second-best.\\
The differences in prediction performance between the baseline methods and MEGAN models can be explained by the slightly different model architectures. However, one particularly interesting result is the small but significant performance difference between $\text{MEGAN}^{1}_{0.0}$ and the supervised $\text{MEGAN}^{1,\text{(S)}}_{0.0}$ version. In both cases, the same model architecture and hyperparameters are used, the only difference being that the latter additionally receives the explanatory information during training. This indicates that the explanations provide the model with some additional level of information about the task, which is useful for the main prediction task as well.\\

\noindent Aside from the numerical results, Figure~\ref{fig:rbmotifs} illustrates one example for these explanations. It shows that the single-explanation methods are able to capture the ground truth explanations to various degrees of success. However, in the presence of motifs with opposing influence, we often observe the issue that single-explanation methods focus on only one of these motifs and fail to highlight the other. An example of this can be seen with the explanation generated by GNNExplainer in Figure~\ref{fig:rbmotifs}, where it only highlights the positive explanation as being important. Although this is not always the case, we believe this effect contributes to the lower explanation accuracy results of these methods. Explanation-supervised training can be used to effectively counter this property, as is evident from the examples and the numerical results. However, even if all the explanatory motifs are correctly highlighted, we argue that single-explanations still don't provide the crucial information about \textit{how} each motif contributes to the prediction outcome, as the polarity information cannot be retrieved from a single channel.


\subsubsection{Multi-Explanations}

To demonstrate the advantages of multi-channel explanations, we conduct an experiment with the RbMotifs dataset, where the ground truth explanatory motifs of each graph are separated into two channels according to their influence on the target values. All blue-based motifs with a negative influence are sorted into one channel and all red-based motifs with positive influence are sorted into another.\\
We train two models to solve the prediction task: A two-channel $\text{MEGAN}^{2}_{1.0}$ model, which uses explanation co-training to promote the generation of explanations according to the previously introduced explanations and a $\text{MEGAN}^{2,\text{(S)}}_{0.0}$ which is explanation-supervised with the ground truth explanations instead. The results can be found in the lower section of Table~\ref{tab:rbmotifs_all}.\\

\noindent Both models achieve nearly equal predictive performance, explanation sparsity, and $\operatorname{Fidelity}^*$. The explanation-supervised model achieves near-perfect explanation accuracy for nodes and edges. However, the explanation co-training model also achieves a very good explanation accuracy. The right-hand side of Figure~\ref{fig:rbmotifs} shows an example of these results. As can be seen, both versions of the model are able to correctly capture the ground truth explanatory motifs according to their respective influence on the target value. The highly positive $\operatorname{Fidelity}^*$ results in both cases prove that both of the model's channels actually contribute to the prediction outcome according to their assigned interpretations of negative and positive influence. The results of this experiment present solid evidence that our proposed explanation co-training is an effective method to accurately capture the polarity of ground truth explanations even in the absence of ground truth explanations during training.



\subsection{Real World Datasets}


\subsubsection{MovieReviews - Sentiment Classification}

The \textit{MovieReviews} dataset is originally a natural language processing dataset from the ERASER benchmark \cite{deyoung_eraser_2020} consisting of 2000 movie reviews from the IMDB database. The general sentiment of each review is labeled as either "positive" or "negative", where both classes are represented equally. Since this is a text classification dataset in its original form, we first process it in a manner similar to Rathee \etal \cite{rathee_bagel_2022}. First, the raw strings are converted into token lists, where tokens are either words or other sentence elements such as punctuation. Each token is converted into a 50-dimensional feature vector through a pre-trained GLOVE model \cite{pennington_glove_2014}. We finally convert the token list into a graph by applying a sliding window method, where each token is considered to be a node and connected to its four closest neighbors through an undirected edge.\\
We train a three-layer $\operatorname{MEGAN}^{2}_{1.0}$ model to solve the binary sentiment classification task for each graph using the classification version of the explanation co-training procedure. The explanation co-training procedure promotes the first explanation channel of the network to contain evidence for the "negative" class label and the second channel for the "positive" class label.\\

\noindent In terms of classification performance our model achieves similar results (F1 $\approx$ 0.85) as previously reported by Rathee \etal \cite{rathee_bagel_2022}, who also use GNN and GLOVE embeddings. However, these results are significantly worse than results obtained with state-of-the-art NLP models, as they are for example reported by DeYoung \etal \cite{deyoung_eraser_2020} (F1 $\approx$ 0.92). We believe the main reason for this difference to be the use of the token embeddings derived from the 2014 GLOVE model. In the future, it would be interesting to see if GNNs could achieve competitive performance by using a state-of-the-art encoder such as BERT \cite{devlin_bert_2019}.\\
In regard to the generated explanations, Table~\ref{tab:movie_review} shows one example of a movie review. As can be seen, the model correctly learns negative adjectives such as "bad" as evidence for the "negative" class and positive adjectives such as "breathtaking" and "best" as evidence for the "positive" class.
Despite this encouraging result, we still find there to be some errors in regard to the model's explanations about sentiment classification. On the one hand, the model also highlights unrelated words as explanations as well, such as "criminal" showing up as an explanation for negative reviews and "director" as positive evidence. On the other hand, the model is also not capable of accurately identifying negations and sarcasm to cause an inversion of sentiment.


\subsubsection{AqSolDB - Molecular Regression}


\begin{table}[t]
    \caption{Example explanations generated for both sentiment classes for a review about the movie "Avengers Endgame". Larger importance values are represented by stronger color highlights.}
    \label{tab:movie_review}
    \definecolor{LightGreen}{rgb}{0.97,1,0.97}
\newcommand{\reducedstrut}{\vrule width 0pt height .9\ht\strutbox depth .9\dp\strutbox\relax}
\newcolumntype{R}[1]{>{\raggedleft\arraybackslash}p{ #1 }}
\newcolumntype{L}[1]{>{\raggedright\arraybackslash}p{ #1 }}
\newcolumntype{Z}[1]{>{\arraybackslash}X{ #1 }}

\setlength\tabcolsep{4pt}

\begin{center}

\sf
\small
\color{darkgray}
\begin{tabular}{ L{5.8cm} L{5.8cm} }
\multicolumn{1}{c}{\color{gray} \small Negative \vspace{2pt} } & 
\multicolumn{1}{c}{\color{gray} \small Positive \vspace{2pt} } \\
\arrayrulecolor{gray}

{\smaller \fontdimen2\font=0em \fboxsep=2pt \colorbox{ red!0 }{\reducedstrut overall } \colorbox{ red!0 }{\reducedstrut avengers } \colorbox{ red!0 }{\reducedstrut endgame } \colorbox{ red!0 }{\reducedstrut was } \colorbox{ red!0 }{\reducedstrut a } \colorbox{ red!0 }{\reducedstrut remarkable } \colorbox{ red!0 }{\reducedstrut movie } \colorbox{ red!0 }{\reducedstrut and } \colorbox{ red!0 }{\reducedstrut a } \colorbox{ red!0 }{\reducedstrut worthy } \colorbox{ red!0 }{\reducedstrut culmination } \colorbox{ red!0 }{\reducedstrut of } \colorbox{ red!0 }{\reducedstrut the } \colorbox{ red!0 }{\reducedstrut mcu } \colorbox{ red!0 }{\reducedstrut up } \colorbox{ red!0 }{\reducedstrut to } \colorbox{ red!0 }{\reducedstrut this } \colorbox{ red!0 }{\reducedstrut point } \colorbox{ red!0 }{\reducedstrut there } \colorbox{ red!0 }{\reducedstrut were } \colorbox{ red!0 }{\reducedstrut some } \colorbox{ red!0 }{\reducedstrut genuinely } \colorbox{ red!0 }{\reducedstrut heartbreaking } \colorbox{ red!0 }{\reducedstrut moments } \colorbox{ red!0 }{\reducedstrut and } \colorbox{ red!0 }{\reducedstrut breathtaking } \colorbox{ red!0 }{\reducedstrut action } \colorbox{ red!0 }{\reducedstrut sequences } \colorbox{ red!0 }{\reducedstrut but } \colorbox{ red!0 }{\reducedstrut to } \colorbox{ red!0 }{\reducedstrut be } \colorbox{ red!3 }{\reducedstrut honest } \colorbox{ red!0 }{\reducedstrut some } \colorbox{ red!3 }{\reducedstrut of } \colorbox{ red!0 }{\reducedstrut the } \colorbox{ red!0 }{\reducedstrut movies } \colorbox{ red!0 }{\reducedstrut i } \colorbox{ red!0 }{\reducedstrut had } \colorbox{ red!0 }{\reducedstrut to } \colorbox{ red!0 }{\reducedstrut sit } \colorbox{ red!0 }{\reducedstrut through } \colorbox{ red!0 }{\reducedstrut to } \colorbox{ red!0 }{\reducedstrut get } \colorbox{ red!0 }{\reducedstrut here } \colorbox{ red!0 }{\reducedstrut were } \colorbox{ red!0 }{\reducedstrut not } \colorbox{ red!0 }{\reducedstrut worth } \colorbox{ red!0 }{\reducedstrut it } \colorbox{ red!0 }{\reducedstrut some } \colorbox{ red!0 }{\reducedstrut of } \colorbox{ red!0 }{\reducedstrut the } \colorbox{ red!0 }{\reducedstrut early } \colorbox{ red!0 }{\reducedstrut mcu } \colorbox{ red!1 }{\reducedstrut movies } \colorbox{ red!2 }{\reducedstrut and } \colorbox{ red!7 }{\reducedstrut series } \colorbox{ red!4 }{\reducedstrut leading } \colorbox{ red!1 }{\reducedstrut up } \colorbox{ red!0 }{\reducedstrut to } \colorbox{ red!0 }{\reducedstrut this } \colorbox{ red!0 }{\reducedstrut finale } \colorbox{ red!0 }{\reducedstrut i } \colorbox{ red!2 }{\reducedstrut found } \colorbox{ red!5 }{\reducedstrut rather } \colorbox{ red!8 }{\reducedstrut bland } \colorbox{ red!10 }{\reducedstrut unfunny } \colorbox{ red!9 }{\reducedstrut and } \colorbox{ red!6 }{\reducedstrut sometimes } \colorbox{ red!6 }{\reducedstrut just } \colorbox{ red!14 }{\reducedstrut downright } \colorbox{ red!46 }{\reducedstrut bad } \colorbox{ red!21 }{\reducedstrut but } \colorbox{ red!47 }{\reducedstrut this } \colorbox{ red!2 }{\reducedstrut movie } \colorbox{ red!3 }{\reducedstrut was } \colorbox{ red!0 }{\reducedstrut one } \colorbox{ red!0 }{\reducedstrut of } \colorbox{ red!0 }{\reducedstrut the } \colorbox{ red!0 }{\reducedstrut best } \colorbox{ red!0 }{\reducedstrut movies } \colorbox{ red!0 }{\reducedstrut i } \colorbox{ red!0 }{\reducedstrut have } \colorbox{ red!0 }{\reducedstrut seen } \colorbox{ red!0 }{\reducedstrut in } \colorbox{ red!0 }{\reducedstrut a } \colorbox{ red!0 }{\reducedstrut while } }
&
{\smaller \fontdimen2\font=0em \fboxsep=2pt \colorbox{ green!3 }{\reducedstrut overall } \colorbox{ green!14 }{\reducedstrut avengers } \colorbox{ green!8 }{\reducedstrut endgame } \colorbox{ green!12 }{\reducedstrut was } \colorbox{ green!11 }{\reducedstrut a } \colorbox{ green!17 }{\reducedstrut remarkable } \colorbox{ green!37 }{\reducedstrut movie } \colorbox{ green!32 }{\reducedstrut and } \colorbox{ green!11 }{\reducedstrut a } \colorbox{ green!11 }{\reducedstrut worthy } \colorbox{ green!0 }{\reducedstrut culmination } \colorbox{ green!0 }{\reducedstrut of } \colorbox{ green!0 }{\reducedstrut the } \colorbox{ green!0 }{\reducedstrut mcu } \colorbox{ green!0 }{\reducedstrut up } \colorbox{ green!0 }{\reducedstrut to } \colorbox{ green!0 }{\reducedstrut this } \colorbox{ green!0 }{\reducedstrut point } \colorbox{ green!0 }{\reducedstrut there } \colorbox{ green!0 }{\reducedstrut were } \colorbox{ green!1 }{\reducedstrut some } \colorbox{ green!15 }{\reducedstrut genuinely } \colorbox{ green!30 }{\reducedstrut heartbreaking } \colorbox{ green!20 }{\reducedstrut moments } \colorbox{ green!51 }{\reducedstrut and } \colorbox{ green!28 }{\reducedstrut breathtaking } \colorbox{ green!12 }{\reducedstrut action } \colorbox{ green!3 }{\reducedstrut sequences } \colorbox{ green!0 }{\reducedstrut but } \colorbox{ green!0 }{\reducedstrut to } \colorbox{ green!0 }{\reducedstrut be } \colorbox{ green!0 }{\reducedstrut honest } \colorbox{ green!0 }{\reducedstrut some } \colorbox{ green!0 }{\reducedstrut of } \colorbox{ green!0 }{\reducedstrut the } \colorbox{ green!0 }{\reducedstrut movies } \colorbox{ green!0 }{\reducedstrut i } \colorbox{ green!0 }{\reducedstrut had } \colorbox{ green!0 }{\reducedstrut to } \colorbox{ green!0 }{\reducedstrut sit } \colorbox{ green!0 }{\reducedstrut through } \colorbox{ green!0 }{\reducedstrut to } \colorbox{ green!0 }{\reducedstrut get } \colorbox{ green!0 }{\reducedstrut here } \colorbox{ green!0 }{\reducedstrut were } \colorbox{ green!0 }{\reducedstrut not } \colorbox{ green!0 }{\reducedstrut worth } \colorbox{ green!0 }{\reducedstrut it } \colorbox{ green!0 }{\reducedstrut some } \colorbox{ green!0 }{\reducedstrut of } \colorbox{ green!0 }{\reducedstrut the } \colorbox{ green!0 }{\reducedstrut early } \colorbox{ green!0 }{\reducedstrut mcu } \colorbox{ green!0 }{\reducedstrut movies } \colorbox{ green!2 }{\reducedstrut and } \colorbox{ green!2 }{\reducedstrut series } \colorbox{ green!1 }{\reducedstrut leading } \colorbox{ green!1 }{\reducedstrut up } \colorbox{ green!0 }{\reducedstrut to } \colorbox{ green!0 }{\reducedstrut this } \colorbox{ green!0 }{\reducedstrut finale } \colorbox{ green!0 }{\reducedstrut i } \colorbox{ green!0 }{\reducedstrut found } \colorbox{ green!0 }{\reducedstrut rather } \colorbox{ green!0 }{\reducedstrut bland } \colorbox{ green!0 }{\reducedstrut unfunny } \colorbox{ green!0 }{\reducedstrut and } \colorbox{ green!0 }{\reducedstrut sometimes } \colorbox{ green!0 }{\reducedstrut just } \colorbox{ green!0 }{\reducedstrut downright } \colorbox{ green!0 }{\reducedstrut bad } \colorbox{ green!0 }{\reducedstrut but } \colorbox{ green!0 }{\reducedstrut this } \colorbox{ green!0 }{\reducedstrut movie } \colorbox{ green!0 }{\reducedstrut was } \colorbox{ green!0 }{\reducedstrut one } \colorbox{ green!2 }{\reducedstrut of } \colorbox{ green!14 }{\reducedstrut the } \colorbox{ green!47 }{\reducedstrut best } \colorbox{ green!80 }{\reducedstrut movies } \colorbox{ green!27 }{\reducedstrut i } \colorbox{ green!11 }{\reducedstrut have } \colorbox{ green!1 }{\reducedstrut seen } \colorbox{ green!0 }{\reducedstrut in } \colorbox{ green!0 }{\reducedstrut a } \colorbox{ green!0 }{\reducedstrut while } }
\\

\end{tabular}
\end{center}
\end{table}

\begin{figure}[t]
    \centering
    \includegraphics[width=\textwidth]{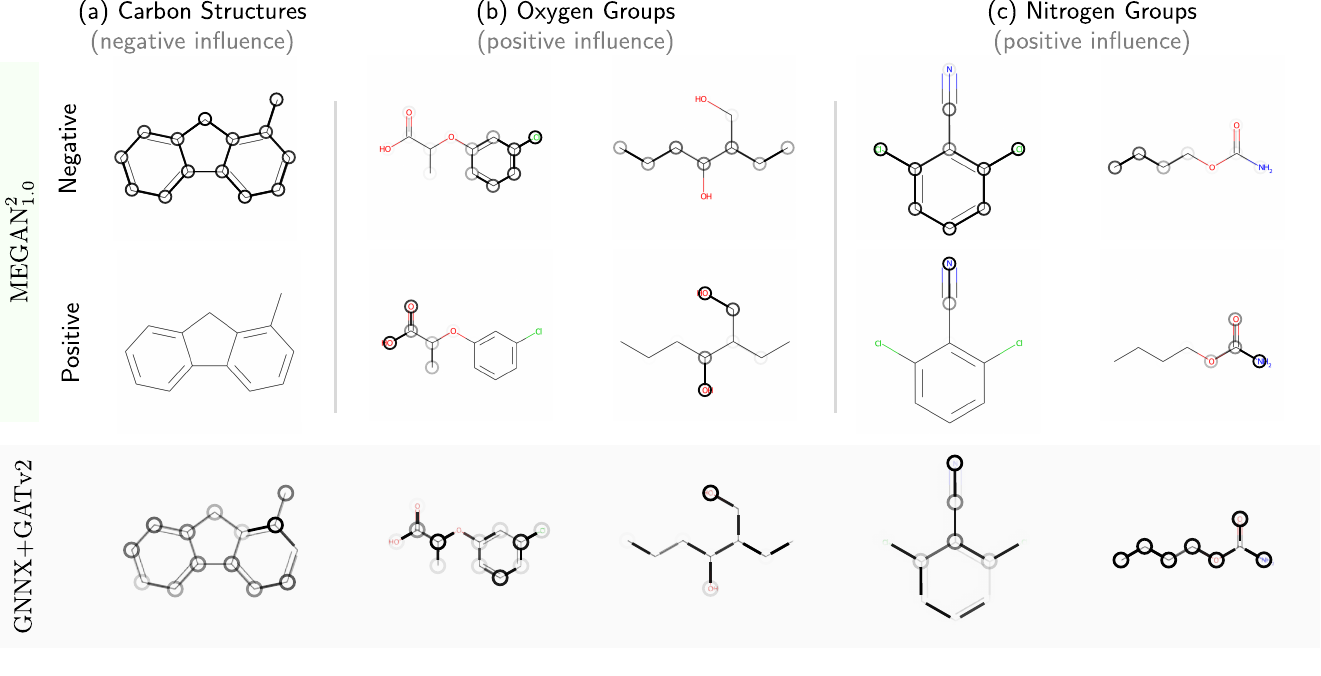}
    \caption{Example explanations generated by MEGAN and GNNExplainer for the prediction of water solubility. Explanations are represented as bold highlights of the corresponding graph elements. Explanations are represented as bold highlights of the corresponding graph elements. (a) Examples of molecules dominated by large carbon structures which are known as negative influences on water solubility. (b) Examples of molecules containing oxygen functional groups which are known to be a positive influence on water solubility. (c) Examples of molecules containing nitrogen groups which are also known as positive influences.}
    \label{fig:solubility}
\end{figure}

\begin{table}[t]
    \caption{Results for 5 independent repetitions of the experiments with the AqSolDB dataset for water solublity. We report the mean in black and the standard deviation in gray.}
    \label{tab:solubility}
    \begin{center}

\definecolor{LightGreen}{rgb}{0.97,1,0.97}
\setlength\tabcolsep{15pt}
\small

\begin{tabular}{lccc}
\toprule
\multicolumn{1}{c}{Model} &
\multicolumn{1}{c}{$R^2 \uparrow$} &
\multicolumn{1}{c}{$\text{Sparsity} \downarrow$} &
\multicolumn{1}{c}{$\text{Fidelity}^{(*)} \uparrow$} \\
\midrule

GNNX+GATv2 &
$0.93 {\color{gray} \pm \mathsmaller{ 0.01 } }$ &
$0.34 {\color{gray} \pm \mathsmaller{ 0.27 } }$ &
$1.26 {\color{gray} \pm \mathsmaller{ 0.90 } }$ \\
\rowcolor{LightGreen}
$\text{MEGAN}^{2}_{1.0}$  & 
$0.93 {\color{gray} \pm \mathsmaller{ 0.01 } }$ &
$0.22 {\color{gray} \pm \mathsmaller{ 0.14 } }$ &
$2.50^{(*)} {\color{gray} \pm \mathsmaller{ 2.29 } }$ \\ 
$\text{Consensus Model}^{\dagger}$ &
$0.93$ &
- &
- \\

\bottomrule
\end{tabular}
\end{center}
    \smaller 
\hspace*{2pt} $^{\dagger}$ Previously published results by Sorkun \etal \cite{sorkun_aqsoldb_2019}. \\
\hspace*{2pt} $^{\text{(*)}}$ Multi-explanation case measures Fidelity$^{*}$ metric
\end{table}

\begin{figure}[t]
    \centering
    \includegraphics[width=\textwidth]{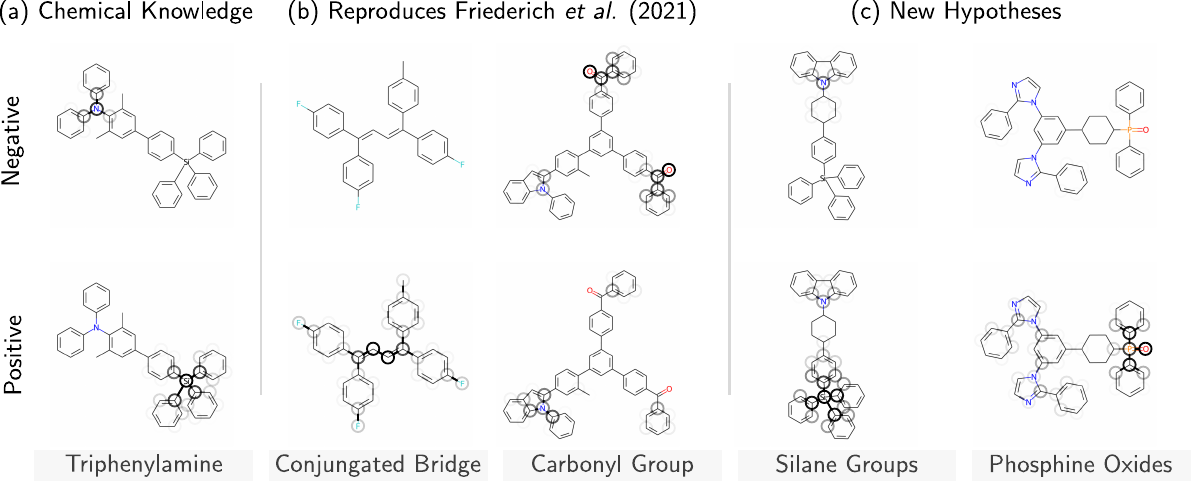}
    \caption{Example explanations obtained from the MEGAN model for the prediction of the singlet-triplet energy gap of the TADF dataset. (a) Explanations that reproduce known chemical intuition about the task. (b) Explanations that reproduce hypotheses previsouly published by Friederich \etal \cite{friederich_scientific_2021}. (c) New explanatory sub-graph motifs proposed through an observation of the explanations generated by MEGAN.}
    \label{fig:tadf}
\end{figure}

The \textit{AqSolDB} \cite{sorkun_aqsoldb_2019} dataset consists of roughly 10000 molecular graphs which are annotated with experimentally determined values of their water solubility. In chemistry, there exists some general intuition about what kinds of molecular structures are responsible for higher solubility values and which are responsible for lower ones. In a simplified manner, one can say that non-polar substructures such as carbon rings and long carbon chains generally result in lower solubility values, while polar structures such as certain nitrogen and oxygen functional groups are associated with higher values.\\
In this experiment, we train a dual-channel three-layer $\operatorname{MEGAN}^{2}_{1.0}$ model to predict the continuous solubility values for the molecular graphs. We make use of the previously described regression version of the co-training procedure, which promotes the first channel to highlight negatively influencing motifs and the second channel to highlight positively influencing motifs. Additionally, we train a comparable GATv2 model on the solubility dataset as well and use GNNExplainer to produce single explanations as a comparison.\\

\noindent Both the MEGAN model and the GATv2 model are able to match the predictive performance which was previously reported in the literature by Sorkunen \etal \cite{sorkun_pushing_2021}. Both approaches also generate explanations with low sparsity and high fidelity values, as it can be seen in Table~\ref{tab:solubility}. Figure~\ref{fig:solubility} illustrates some example explanations generated by MEGAN and GNNExplainer. The examples show that the explanations generated by MEGAN match the general human intuition about the structure-property relationships of water solubility. Large carbon structures are consistently highlighted in the negative explanation channel. The positive explanation channel on the other hand mostly contains polar nitrogen and oxygen functional groups. The explanations generated by GNNExplainer on the basis of the GATv2 model, however, do not show any such discernible pattern.\\

\noindent Despite an equally high predictive performance and high explanation fidelity, we argue that the single-explanation case contributes significantly less useful information for a human understanding of the predictions. We think this example reinforces the importance of the multi-explanation approach, especially for graph regression problems. By considering the polarity of structure-property explanations in graph regression problems, the MEGAN model is able to provide explanations that are more consistent with human intuition and are thus more interpretable.


\subsubsection{TADF - Molecular Regression}

Previous experiments were able to provide exemplary evidence for the correctness of MEGAN's explanations through real-world datasets for which human intuition exists. In this final experiment, we choose a dataset where almost no human intuition exists to investigate potential applications to reveal novel insights about structure-property relationships.\\
The \textit{TADF} dataset consists of roughly half a million molecular graphs. Target value annotations were during a high-throughput virtual screening experiment conducted by Gómez-Bombarelli \etal \cite{gomez-bombarelli_design_2016} with the objective to discover novel materials for an application in OLED technology. Specifically, the authors aimed to discover materials that show a specific characteristic of thermally delayed fluorescence (TADF). This class of materials is a promising approach to avoid the high cost of typically used phosphorescent OLED materials \cite{endo_efficient_2011,zhang_design_2012}. Along the delayed fluorescent rate constant $k_{\text{TADF}}$, the elements of the dataset are annotated with the singlet-triplet energy gap $\Delta E_{\text{st}}$ and the oscillator strength $f$.\\
In this experiment, we train a three-layer $\operatorname{MEGAN}^{2}_{1.0}$ model to estimate the singlet-triplet gap $\Delta E_{\text{st}}$ for each element. As before, the explanation co-training promotes the first channel to contain the negative influences and the second channel to contain the positive influences.\\

\noindent Our model achieves overall good predictivity ($\text{R}^{2} \approx 0.90$) for the main prediction task and a positive $\operatorname{Fidelity}^*$ value validating that the individual channels indeed affect the model prediction according to their pre-determined interpretations. Figure~\ref{fig:tadf} illustrates some example explanations obtained from the model. Most importantly, we show that our model is able to replicate one of the few known structure-property relationships about the singlet-triplet energy. Triphenylamine bridges are known to be associated with low energy gaps, as they cause the necessary twist angles between the fragments, decoupling electron-donating and electron-accepting parts of a molecule to reduce the exchange interaction between the frontier orbitals which would otherwise lower the triplet state compared to the singlet state, thus preventing undesired singlet-triplet splittings. This fact is reflected in Figure~\ref{fig:tadf}(a), where a triphenylamine bridge is highlighted as a negative influence on the prediction outcome. Furthermore, our model is able to support hypotheses published in previous work by Friederich \etal \cite{friederich_scientific_2021}, who use an interpretable decision tree method to generate explanation hypotheses for the same task. As shown in Figure~\ref{fig:tadf}(b) our model replicates their findings of conjugated bridges as a positive influence on the energy gap and carbonyl groups as a negative influence. Beyond that, our model finds several novel hypotheses about structure-property relationships, two of which are shown in Figure~\ref{fig:tadf}(c): We can propose silane groups and phosphine oxides as positive influences to the singlet-triplet energy gap.

\section{Limitations}

Despite the encouraging experimental results, there are limitations to the proposed MEGAN architecture: Firstly, there is no hard guarantee that each channel's explanations align correctly according to their pre-determined interpretations. This alignment is mainly promoted through the explanation co-training, whose influence on the network is dependent on a hyperparameter. We occasionally observed "explanation leakage" and "explanation flipping" during training. In those rare cases, explanations factually belonging to one channel may either faintly appear in the opposite channel or a particularly disadvantageous initialization of the network causes explanations to develop in the exact opposite channel relative to their assigned interpretation. Ultimately, the alignment of a particular channel with its intended interpretation has to be tested through a Fidelity* analysis after the model training.\\
The second limitation is in the design of the explanation co-training itself, which essentially reduces the problem to a subgraph counting task. While there are many important real-world applications that can be approximated as such, it still presents an important limit to the expressiveness of the models produced by our model.

\section{Conclusion}

In this work, we introduce the self-explaining multi-explanation graph attention network (MEGAN) architecture, which produces node and edge attributional explanations for graph regression and classification tasks. Our model implements the number $K$ of generated explanations as a hyperparameter of the network itself, instead of being dependent on the task specification. Based on several examplery synthetic and real-world datasets, we show that this property is especially crucial for graph regression problems. By being able to generate attributional explanations for a single regression target along multiple explanation channels, our model is able to account for the \textit{polarity} of explanations. In many graph regression applications certain sub-graph motifs influence the predicted outcome in opposing directions: Some motifs present a negative influence on the overall prediction, while others are a positive influence. We achieve the alignment of the model's multiple explanation channels according to these pre-determined interpretations by introducing an explanation co-training procedure. Beside the main prediction loss, an additional explanation loss is generated from an approximate solution of the prediction problem based only on each channels explanation masks. We can validate the channel's alignment to their respective intended interpretations through the $\operatorname{Fidelity}^{*}$ metric, which extends the concept of explanation fidelity to our multi-channel case.\\
Additionally, we demonstrate the capabilities of our model for explanation-supervised training, where a model is trained to produce explanations based on a set of given ground truth explanations. For a synthetic graph regression dataset, we show that our model is able to learn the given ground truth explanations almost perfectly, significantly outperforming an existing baseline method from literature.\\

\noindent One particularly interesting result is the improvement of the prediction performance for the explanation-supervised training during the first synthetic experiment but not during the second one. Similar effects have already been shown in the domain of image processing, where various authors are able to demonstrate a performance increase when models are additionally trained to emulate human saliency maps \cite{linsley_learning_2019,boyd_cyborg_2022}. One promising direction for future work will be to investigate the conditions under which (human) explanations have the potential to improve predictive performance for graph-related tasks as well. 
 

\section{Reproducibility Statement}

We make our experimental code publically available at \url{https://github.com/aimat-lab/graph_attention_student}. The code is implemented in the Python 3.9 programming language. Our neural networks are built with the KGCNN library by Reiser \etal \cite{reiser_graph_2021}, which provides a framework for graph neural network implementations with TensorFlow and Keras. We make all data used in our experiments publically available on a file share provider \url{https://bwsyncandshare.kit.edu/s/E3MynrfQsLAHzJC}. The datasets can be loaded, processed, and visualized with the visual graph datasets package \url{https://github.com/aimat-lab/visual_graph_datasets}. All experiments were performed on a system with the following specifications: Ubuntu 22.04 operating system, Ryzen 9 5900 processor, RTX 2060 graphics card and 80GB of memory. We have aimed to package the various experiments as independent modules and our code repository contains a brief explanation of how these can be executed.

%
\clearpage
\bibliographystyle{splncs04}
\bibliography{megan}

\noindent
\appendix

\section{Evaluation Metrics}


\subsubsection{Fidelity}
\label{app:fidelity}

Fidelity metrics are used to quantify the degree to which explanations are actually responsible for a model's prediction. In our experiments, we use the definition of the $\operatorname{Fidelity}^+$ metric as defined by Yuan \etal \cite{yuan_explainability_2022}. It is calculated as the difference between the original predicted value and the predicted value if the elements of the explanation are removed from the input graph. It is generally assumed the higher this value, the more important those elements are for the prediction. This metric generally works well by itself for classification problems, where confidence values are limited to the range between 0 and 1. In such a case, a fidelity value of 0.8 would be considered quite high because there exists a frame of reference that defines 1 as the maximum possible value. However, for this reason, we find that the metric is not immediately applicable to the regression problems since there exists no frame of reference as to what would be considered a particularly high or low value.\\
Instead, for our regression experiments, we use a relative fidelity value which is defined relative to a point of reference. 
\begin{equation}
    \operatorname{Fidelity}^+_{\text{rel}} = \operatorname{Fidelity}^+ - \operatorname{Fidelity}^+_{\text{random}}
\end{equation}
As the frame of reference, we use the fidelity value which results from a purely random input graph mask, which has the \textit{same sparsity} as the given explanation. The random fidelity value is calculated as the arithmetic mean resulting from 10 such randomly sampled input masks per explanation.


\section{GNES Implementation}

In our experiments, we use the GNES method by Gao \etal \cite{gao_gnes_2021} as a baseline approach from the literature that supports explanation supervision. In their framework, the authors propose using existing differentiable post-hoc explanation methods for explanation supervision. For that, they introduce a generic framework to describe node and edge attributional explanations. For example, they define node the attributional explanation for node $n$ at layer $l$ as 
\begin{equation}
    M_n^{(l)} = ||\; \text{ReLU}( g(\frac{\partial y_c}{\partial F_n^{(l)}}) \cdot h(F_n^{(l)}) ) \;||
\end{equation}
where $F_n^{(l)}$ is the activation of node $n$ at layer $l$. $g(\cdot)$ and $h(\cdot)$ are generic functions that can be defined for specific implementations of explanation methods. Edge explanations are defined in a similarly generic way. Explanation supervision is then achieved through additional loss MAE loss terms between these generated explanations and the given reference explanations.\\

\noindent For our experiments, we were not able to use the original code at \url{https://github.com/YuyangGao/GNES} as that implementation only supports binary classification problems and is limited to a batch size of 1. We re-implement their method in the KGCNN framework. We follow the original paper as closely as possible for the version we call $\text{GNES}_\text{original}$. However, we find that the used $\text{ReLU}(\cdot)$ operation does not work well with regression operations as it cuts off negative values and thus actively discards explanatory motifs with \textit{opposing influence}. Consequently, we modify the method to use an absolute value operation $|| \cdot ||$ instead of the $\text{ReLU}(\cdot)$ for the version we call $\text{GNES}_{\text{fixed}}$. We find that this version works much better with regression tasks as it is able to properly account for positive and negative influences.


\section{GNN Benchmarks}

Aside from its capability for explanation supervision, we also find that our model generally shows a good prediction performance as well, when compared to other state-of-the-art GNNs. Figure~\ref{fig:benchmarks} shows the benchmarking results of the MEGAN model compared to several other GNNs from the literature for two datasets of molecular property prediction. The benchmarking results were obtained from the KGCNN library \url{https://github.com/aimat-lab/gcnn_keras/tree/master/training/results}. To produce the results, all models were subjected to a cursory hyperparameter optimization on the respective datasets. The MEGAN models trained for this comparison use neither explanation supervision nor the co-training method.\\
The results show that MEGAN achieves the second-best results for both tasks. 


\begin{figure}[b!]
\centering
    
\begin{subtable}[t]{0.48\textwidth}
    \caption{Results for the ESOL dataset \cite{delaney_esol_2004} which consists of 1128 molecular graphs and their respective values for water solubility.}
    \label{tab:esol}
    \setlength\tabcolsep{5pt}
\definecolor{LightGreen}{rgb}{0.97,1,0.97}

\begin{center}
\begin{tabular}{ lcc }
\toprule

\multicolumn{1}{c}{ Model }  & 
\multicolumn{1}{c}{ MAE $\downarrow$}  & 
\multicolumn{1}{c}{ RMSE $\downarrow$}  \\ \midrule





GAT  & 
$0.49{ \color{gray} \mathsmaller{ \pm 0.02 } } $  & 
$0.70{ \color{gray} \mathsmaller{ \pm 0.04 } } $  \\ 

GIN  & 
$0.50{ \color{gray} \mathsmaller{ \pm 0.02 } } $  & 
$0.70{ \color{gray} \mathsmaller{ \pm 0.03 } } $  \\ 

CMPNN  & 
$0.48{ \color{gray} \mathsmaller{ \pm 0.03 } } $  & 
$0.68{ \color{gray} \mathsmaller{ \pm 0.02 } } $  \\ 

INorp  & 
$0.49{ \color{gray} \mathsmaller{ \pm 0.01 } } $  & 
$0.68{ \color{gray} \mathsmaller{ \pm 0.03 } } $  \\ 

GATv2  & 
$0.47{ \color{gray} \mathsmaller{ \pm 0.03 } } $  & 
$0.67{ \color{gray} \mathsmaller{ \pm 0.03 } } $  \\ 

Schnet  & 
$0.46{ \color{gray} \mathsmaller{ \pm 0.03 } } $  & 
$0.65{ \color{gray} \mathsmaller{ \pm 0.04 } } $  \\ 


DMPNN  & 
$0.45{ \color{gray} \mathsmaller{ \pm 0.02 } } $  & 
$0.63{ \color{gray} \mathsmaller{ \pm 0.02 } } $  \\ 

AttentiveFP  & 
$0.46{ \color{gray} \mathsmaller{ \pm 0.01 } } $  & 
$0.63{ \color{gray} \mathsmaller{ \pm 0.03 } } $  \\ 

\rowcolor{LightGreen}
MEGAN  & 
$\underline{0.44}{ \color{gray} \mathsmaller{ \pm 0.03 } } $  & 
$\underline{0.60}{ \color{gray} \mathsmaller{ \pm 0.05 } } $  \\ 

PAiNN  & 
$\mathbf{0.43}{ \color{gray} \mathsmaller{ \pm 0.02 } } $  & 
$\mathbf{0.60}{ \color{gray} \mathsmaller{ \pm 0.02 } } $  \\

\end{tabular}
\end{center}
\end{subtable}
\hfill
\begin{subtable}[t]{0.48\textwidth}
    \caption{Results for the LIPOP dataset \cite{wu_moleculenet_2018} which consists of 4200 molecular graphs and their respective octanol/water distribution coefficient.}
    \label{tab:lipop}
    \setlength\tabcolsep{5pt}
\definecolor{LightGreen}{rgb}{0.97,1,0.97}

\begin{center}
\begin{tabular}{ lcc }
\toprule

\multicolumn{1}{c}{ Model }  & 
\multicolumn{1}{c}{ MAE $\downarrow$}  & 
\multicolumn{1}{c}{ RMSE $\downarrow$}  \\ \midrule
GAT  & 
$0.50{ \color{gray} \mathsmaller{ \pm 0.02 } } $  & 
$0.70{ \color{gray} \mathsmaller{ \pm 0.04 } } $  \\ 

INorp  & 
$0.46{ \color{gray} \mathsmaller{ \pm 0.01 } } $  & 
$0.65{ \color{gray} \mathsmaller{ \pm 0.01 } } $  \\ 

Schnet  & 
$0.48{ \color{gray} \mathsmaller{ \pm 0.00 } } $  & 
$0.65{ \color{gray} \mathsmaller{ \pm 0.00 } } $  \\ 

GIN  & 
$0.45{ \color{gray} \mathsmaller{ \pm 0.01 } } $  & 
$0.64{ \color{gray} \mathsmaller{ \pm 0.03 } } $  \\ 


AttentiveFP  & 
$0.45{ \color{gray} \mathsmaller{ \pm 0.01 } } $  & 
$0.62{ \color{gray} \mathsmaller{ \pm 0.01 } } $  \\ 

GATv2  & 
$0.41{ \color{gray} \mathsmaller{ \pm 0.01 } } $  & 
$0.59{ \color{gray} \mathsmaller{ \pm 0.01 } } $  \\ 

PAiNN  & 
$0.40{ \color{gray} \mathsmaller{ \pm 0.01 } } $  & 
$0.58{ \color{gray} \mathsmaller{ \pm 0.03 } } $  \\ 

CMPNN  & 
$0.41{ \color{gray} \mathsmaller{ \pm 0.01 } } $  & 
$0.58{ \color{gray} \mathsmaller{ \pm 0.01 } } $  \\ 

\rowcolor{LightGreen}
MEGAN  & 
$\underline{0.40}{ \color{gray} \mathsmaller{ \pm 0.01 } } $  & 
$\underline{0.56}{ \color{gray} \mathsmaller{ \pm 0.01 } } $  \\ 

DMPNN  & 
$\mathbf{0.38}{ \color{gray} \mathsmaller{ \pm 0.01 } } $  & 
$\mathbf{0.55}{ \color{gray} \mathsmaller{ \pm 0.03 } } $  \\

\end{tabular}
\end{center}
\end{subtable}

\caption{Benchmarking results obtained from the KGCNN library from a random 5-fold cross-validation.}
\label{fig:benchmarks}
    
\end{figure}

\end{document}